\documentclass[runningheads]{llncs}
\usepackage{graphicx}
\usepackage{amsmath,amssymb} 
\usepackage{color}
\usepackage[width=122mm,left=12mm,paperwidth=146mm,height=193mm,top=12mm,paperheight=217mm]{geometry}

\usepackage{textcomp} 
\usepackage[linesnumbered,ruled,vlined]{algorithm2e}
\usepackage{subcaption}
\usepackage[compatibility=false, font=footnotesize,labelfont=bf]{caption}
\usepackage{booktabs}

\usepackage{hyperref}

\begin{document}

\pagestyle{headings}
\mainmatter

\title{A Discriminative Framework for Anomaly Detection in Large Videos}

\titlerunning{A Discriminative Framework for Anomaly Detection in Large Videos}

\authorrunning{Allison Del Giorno, J. Andrew Bagnell, Martial Hebert}

\author{Allison Del Giorno, J. Andrew Bagnell, Martial Hebert}

\institute{Carnegie Mellon University}

\maketitle

\begin{abstract}
We address an anomaly detection setting in which training sequences are unavailable and anomalies are scored independently of temporal ordering.  Current algorithms in anomaly detection are based on the classical density estimation approach of learning high-dimensional models and finding low-probability events. These algorithms are sensitive to the order in which anomalies appear and require either training data or early context assumptions that do not hold for longer, more complex videos. By defining anomalies as examples that can be \textit{distinguished} from other examples in the same video, our definition inspires a shift in approaches from classical density estimation to simple discriminative learning. Our contributions include a novel framework for anomaly detection that is (1) independent of temporal ordering of anomalies, and (2) unsupervised, requiring no separate training sequences. We show that our algorithm can achieve state-of-the-art results even when we adjust the setting by removing training sequences from standard datasets.
\end{abstract}

\section{Introduction}

\begin{figure*}[t]
\centering
\includegraphics[width=0.8\textwidth,keepaspectratio]{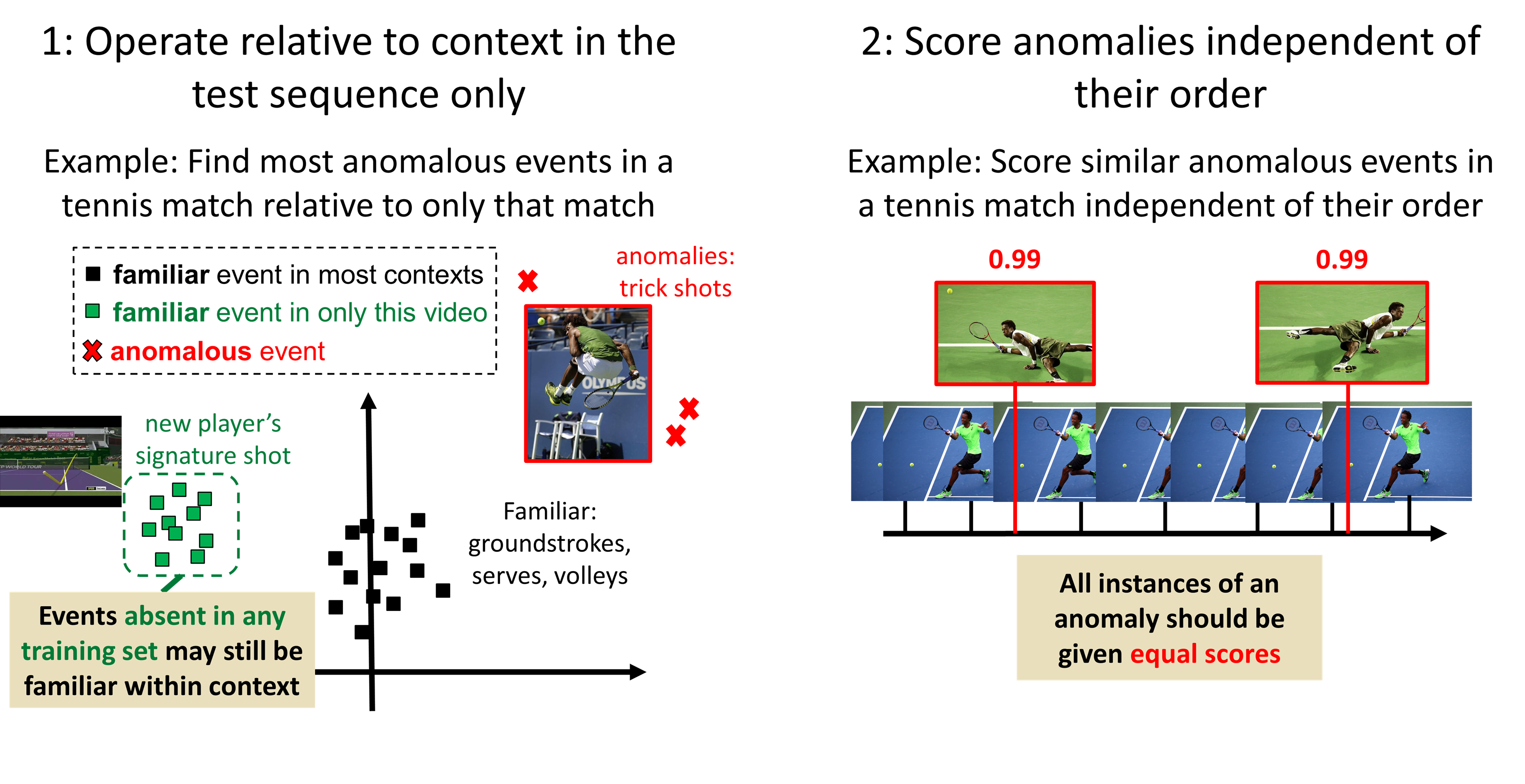}
\vspace{-4mm}
\caption{  
\textbf{Characteristics of our anomaly detection setting. } \textbf{Left: No training sequences. }This setting occurs when we want context to be drawn solely from the test video (e.g. - the player's shot distribution differs for each opponent, and we want anomalies relative to how they played that opponent), or unavailable (a player with a new style debuts at a tournament).  \textbf{Right: Temporal independence. } Often we want to find the most anomalous frames regardless of the order they appear in.
\vspace{-2mm}
} \label{fig:challenges}
\end{figure*}

Anomaly detection is an especially challenging problem because, while its applications are prevalent, it remains ill-defined.  Where there have been attempts at definitions, they are often informal and vary across communities and applications.  In this paper, we define and propose a solution for a largely neglected subproblem within anomaly detection, where two constraints exist: (1) no additional training sequences are available; (2) the order in which anomalies occur should not affect the algorithm's performance on each instance (Figure~\ref{fig:challenges}).  This is an especially challenging setting because we cannot build a model in advance and find deviations from it; much like clustering or outlier detection, the context is defined by the video itself.  This setting is prominent in application fields such as robotics, medicine, entertainment, and data mining.  For instance:

\begin{itemize}
    \item \textit{First-time data.} A robotics team wants to create a robust set of algorithms.  They teleoperate a robot performing a new task or operating in a new environment. The team would like to find out what special cases the robot may have to handle on the perception side, so they ask for a list of the most anomalous instances according to the robot's sensor data relative to that day's conditions and performance.
    \item \textit{Personalized results: context semantically defined as coming only from the test set.} (a) A father wants to find the most interesting parts of the 4-hour home video of his family's Christmas. (b) A healthcare professional wants to review the most anomalous footage of an elderly patient while living under at-home nursing care over the past week.
    \item \textit{Database sifting.} A consulting analyst is told to find abnormal behavior in a large amount of video from a surveillance camera.
\end{itemize}
These illustrate just a few of the practical cases in which it is important to identify all instances of anomalies, regardless of the order in which they appear, and to do so within the context of the testing video.  If videos are available for providing context beyond the test video or it is acceptable to ignore later anomalies as long as the first instance is recognized, there are many mature methods that apply (see~\ref{section:relatedwork}). There are also natural extensions to our method that could incorporate additional context if it were available.  Here we consider a challenging setting in which the context must be derived from the test video and the order in which anomalies occur does not affect their score.

In general anomaly detection settings, one cannot use traditional supervised approaches because it is impossible to find a sufficiently representative set of anomalies.  In our setting, we are given no context ahead of time; unlike other algorithms, we cannot even build a distribution for a representative set of familiar events.  We require the use of approaches that operate solely on the test sequence and adapt to each video's context.  This leads us to denote frames as \textit{anomalous} if they are easily distinguished from other frames in the same video, and \textit{familiar} otherwise.

\subsection{Previous approaches} \label{section:relatedwork}
Anomaly detection presents a set of unique challenges beyond those seen in the supervised learning paradigm. The inability to use training data for both classes of data (familiar and anomalous) leads to two possible approaches: (1) Estimate a model distribution of the familiar and then classify sufficiently large deviations as anomalous; or (2) Seek out points that are identifiable in the distribution of all frames and label those anomalous.  While these approaches seem similar on the surface, they lead to distinct methodologies that differ in the assumptions and data required as well as the type of anomalies they identify.  We show that the latter will satisfy our setting while the former will not, and comes with a few other advantages.

The traditional approach for anomaly detection involves learning a model of familiarity for a given video.  Subsequent time points can then be identified as anomalous if they deviate from the model by some distance metric or inverse likelihood.  We call this set of approaches ``scanning'' techniques.  Examples in this area include sparse reconstruction \cite{Zhao,lu2013abnormal}, dynamic textures \cite{mahadevan2010anomaly,kim2009observe,li2014anomaly}, and human behavior models \cite{mehran2009abnormal}.  The methods take a set of training data, either from separate videos or from hand-chosen frames at the beginning of the video, and build a model.  Many of these methods update models online, and while some do not need to update the model in temporal order~\cite{antic2011video}, they still need a large amount of training data for initialization.   One method in particular achieves reasonable performance with a small number of starting frames~\cite{roshtkhari2013online}, but still requires manual identification of these frames for the algorithm. Generative models work well for domains in which the assumed model of normalcy fits the data well.  This applies when the model is complex enough to handle a variety of events, or when known context can allow the learner to anticipate a model that will fit the data.  These methods generally assume that the features come from a predetermined type of distribution and therefore are likely to fail if the feature distribution changes.  For complex models, computational complexity and the amount of `normal' training data needed to initialize the model becomes a significant bottleneck. Parameter choices also have a larger effect on the ability of the algorithm to fit the data.

Scanning approaches do not satisfy our anomaly detection setting because they violate the two conditions we specified: (1) they require training instances, and (2) they depend on temporal ordering.  Building a model in temporal order of the video makes strong assumptions about the way anomalies should be detected.  In our setting, we must find the `most anomalous' events in the video, regardless of their order.  By building models with updates in temporal order, events that occur earlier in the video are more likely to be anomalous.  For instance, with an event type that occurs only twice in a large video, the first instance will be detected once but the second instance will be ignored.  By choosing a discriminative algorithm that acts independently of the ordering of the video, we avoid these assumptions and pitfalls of the scanning techniques.

Our method shares the discriminative spirit of previous works using saliency for anomaly detection~\cite{mahadevan2010anomaly,li2014anomaly}.  However, the saliency methods used require training data to run and only use local context.  Our objective is to obtain a fully unsupervised method that uses the context of the entire video and is independent of the ordering in which the anomalies occur. \cite{calderara2011detecting} builds a graph and finds anomalies independent of their ordering.  However, it is model-based and only designed to work with trajectories; our goal is discriminative and able to operate on any set of features.

The primary challenge in our setting is our inability to assume the form of the underlying distribution.  A non-parametric method is preferable so that it can generalize to many domains with few assumptions. Permutation tests are nonparametric methods designed to handle such cases.  The general idea is to test the fidelity of a given statistic against a set of other possible statistics from a differently-labeled dataset.  We use a similar approach to test the distinctiveness of each frame.  In our method, the analogous statistic is the ease with which a given data point can be distinguished from other points sampled from the same video.  By testing a frame's distinguishability from different groups of frames, we form a more accurate picture of its global anomaly score.

\subsection{Our Approach}
Our approach is to directly estimate the discriminability of frames with reference to the context in the video.  We do not need a model of every normal event to generate scores for anomalous frames; we can simply attempt to discriminate between anomalous frames and familiar frames to see if there is a difference in the distributions.  We present a framework that tests the discriminability of frames.  In this framework, we perform change detection on a sequence of data from the video to see which frames are distinguishable from previous frames.  Because we want these comparisons to be independent of time, we create \textit{shuffles} of the data by permuting the frames before running each instance of the change detection.  Simple classifiers are used to avoid overfitting and so that each frame will be compared against many permutations of other frames.  This discriminative framework allows us to perform in a unique setting. 

Our contributions are as follows:
\begin{itemize}
\item A permutation-based framework for anomaly detection in a setting free from training data and independent of ordering,
\item A theory that guides the choice of key parameters and justifies our framework,
\item Experimental evaluation demonstrating that this technique achieves performance similar to other techniques that require training data
\end{itemize}

\subsection{A Motivating Example}

\begin{figure}
\includegraphics[width=\linewidth,keepaspectratio,trim={2.5cm 0 2.5cm 0},clip]{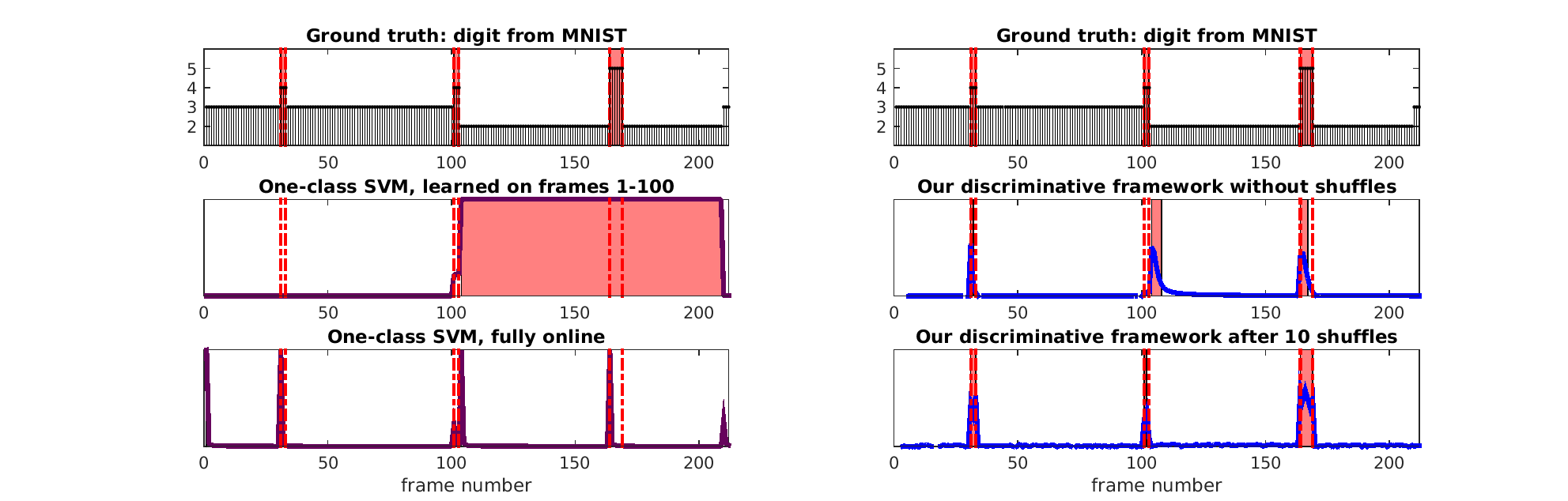}
\caption{  
\textbf{Detections from one-class SVM and our algorithm on a toy example.}  The ground truth represents the digit classes from MNIST that were used to generate each frame.  The red dashes indicate locations of the anomalies.  The red shaded region represents detections made by each algorithm.  Our algorithm without shuffling has the same temporal disadvantages as online one-class SVM.  By including shuffling, we do not trigger false positives on prevalent examples when seen for the first time.  We also detect the full extent of each anomaly and avoid assuming the beginning is familiar.\vspace{-2mm}} \label{fig:oneclassMNIST}
\end{figure}

To motivate our method and demonstrate its advantages over scanning techniques, let's walk through a toy example.  Suppose we draw four images from the MNIST dataset, each with a different label (2, 3, 4, or 5). Then we create a `video' using noisy copies of these images.  The order of these images is shown in Figure~\ref{fig:oneclassMNIST}.  While the first portion of the video contains only instances of `3', both the 3's and 2's are prevalent.  In this case, we would hope that the algorithm classifies all instances of 4's and 5's as anomalous and considers all 2's and 3's familiar.  We use one-class SVM with a RBF kernel as an instance of scanning techniques.  Figure~\ref{fig:oneclassMNIST} shows scores from a static one-class SVM trained on the first portion of the video, the same algorithm with an online update, and our algorithm with and without shuffling.  Our algorithm's performance without shuffling is similar to that of the online one-class SVM.  When the model remains static after the first third of the video, all of the 2's are classified as anomalous.  Even with an online model update, the first few 2's are classified as anomalous.  In addition, not all of the 4's and 5's are given equal anomaly weights within their respective classes.  Our algorithm avoids these pitfalls once shuffling is introduced, classifying only the 4's and 5's as anomalous.  By using a permutation-based framework, we are able to evade assumptions of familiarity and remove the effects of temporal ordering on anomaly scores.

The issues discussed extend beyond just this toy example.  With scanning methods, anomalies that appear more than once may be missed.  In addition, it is common to see failures due to the assumption that the beginning of the video represents familiarity, both by anomalies appearing in the beginning and by other familiar events appearing later in the video.  For videos where context changes frequently (imagine a concert light show whose theme changes every song), this can create a dangerously high number of false positives.  Also note that while this example uses one-class SVM as an example, all scanning techniques have the same inherent problems.  Our method was developed in part to circumvent these previously unavoidable failure cases.  In addition, we hope to demonstrate that simple discriminative techniques can match the performance of more complex generative methods while operating in the new setting we have identified.

\section{Method}
\textbf{Taking the direct approach.} Inspired by density ratio estimation for change point detection~\cite{Liu2013,ito2012,kawahara2009change}, we take a more direct approach to anomaly detection than the popular generative approach.  The main objective of density ratio estimation is to avoid doing unnecessary work when deciding from which one of two distributions a data point was generated: rather than model both distributions independently, we can directly compute the ratio of probabilities that a data point is drawn from one or the other.  This shortcut is especially helpful in anomaly detection.  We are more interested in the \textit{relative} probability that a given frame is anomalous rather than familiar, and are less interested in the distribution of familiar events.  The machine learning community has covered several ways to estimate this ratio directly and has enumerated the several cost functions and other paradigms in which this ratio appears~\cite{sugiyama2010density}.  We note that one such way to estimate these ratios directly is simple logistic regression, and therefore we use this standard classifier as a measure of the deviation between two groups of points (see ``Larger window sizes decrease the effect of overfitting classifiers'' in Section~\ref{sec:theorycomplexity} for formal justification).

\begin{figure*}[t]
\centering
\includegraphics[width=1\textwidth,trim={0 2cm 0 0},clip]{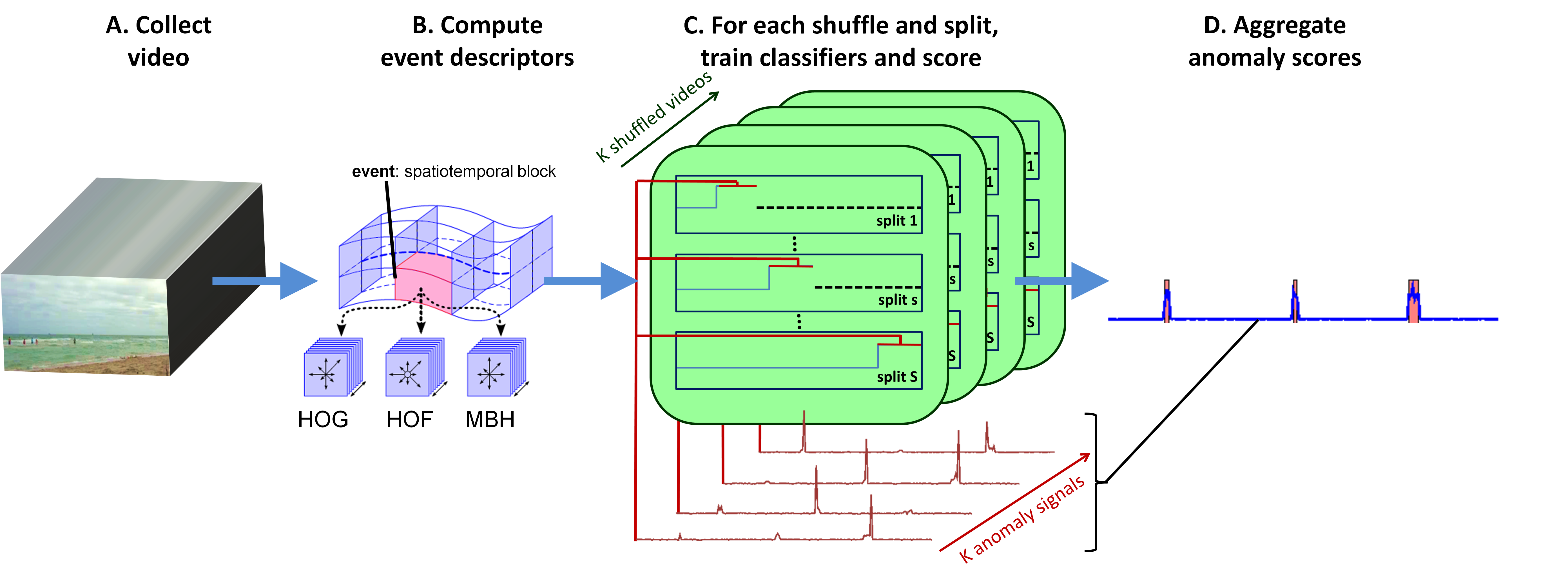}
\caption{
\textbf{The proposed anomaly detection framework.} (A) Given an input video, (B) a descriptor for each frame is passed into the anomaly detection algorithm, (C) where the descriptors are shuffled $K$ times. For each shuffle, the algorithm evaluates anomaly scores for a sliding window of frames. This score is based on the density ratio compared to frames that came before the sliding window. (D) Finally, the scores are combined with averaging to produce the final output signal. Image depicting dense trajectory features is from Wang et al\cite{wang2011action}.
\vspace{-2mm}
} \label{fig:framework}
\vspace{-2mm}
\end{figure*}

\textbf{System overview.} The full framework is depicted in Figure~\ref{fig:framework}.  Recall our definition of anomalous frames: those that are easily distinguished from others in the same video.  Because this definition avoids domain-specific notions of anomalies, it relies on a robust set of features that can be used to distinguish anomalies in a variety of domains.  We assume an appropriate set of features has been computed and forms a descriptor for each frame.  Because this is a discriminative method, the choice of features has a smaller impact on the choice of algorithm and parameters than it would for a generative method.  The overall proposed framework is agnostic to the feature choice; the user can plug in any relevant or state-of-the-art features based on domain knowledge or novel feature methods.  In addition, features can be aggregated within or across frames to obtain different levels of spatial or temporal resolution.  Because the framework does not make explicit assumptions about the distribution of the features, these are simply design choices based on the cost of feature computation and the desired resolution of detections.

\begin{figure}[t]
\centering
\begin{minipage}[t]{.55\textwidth}
    \centering
    \includegraphics[width=0.9\linewidth]{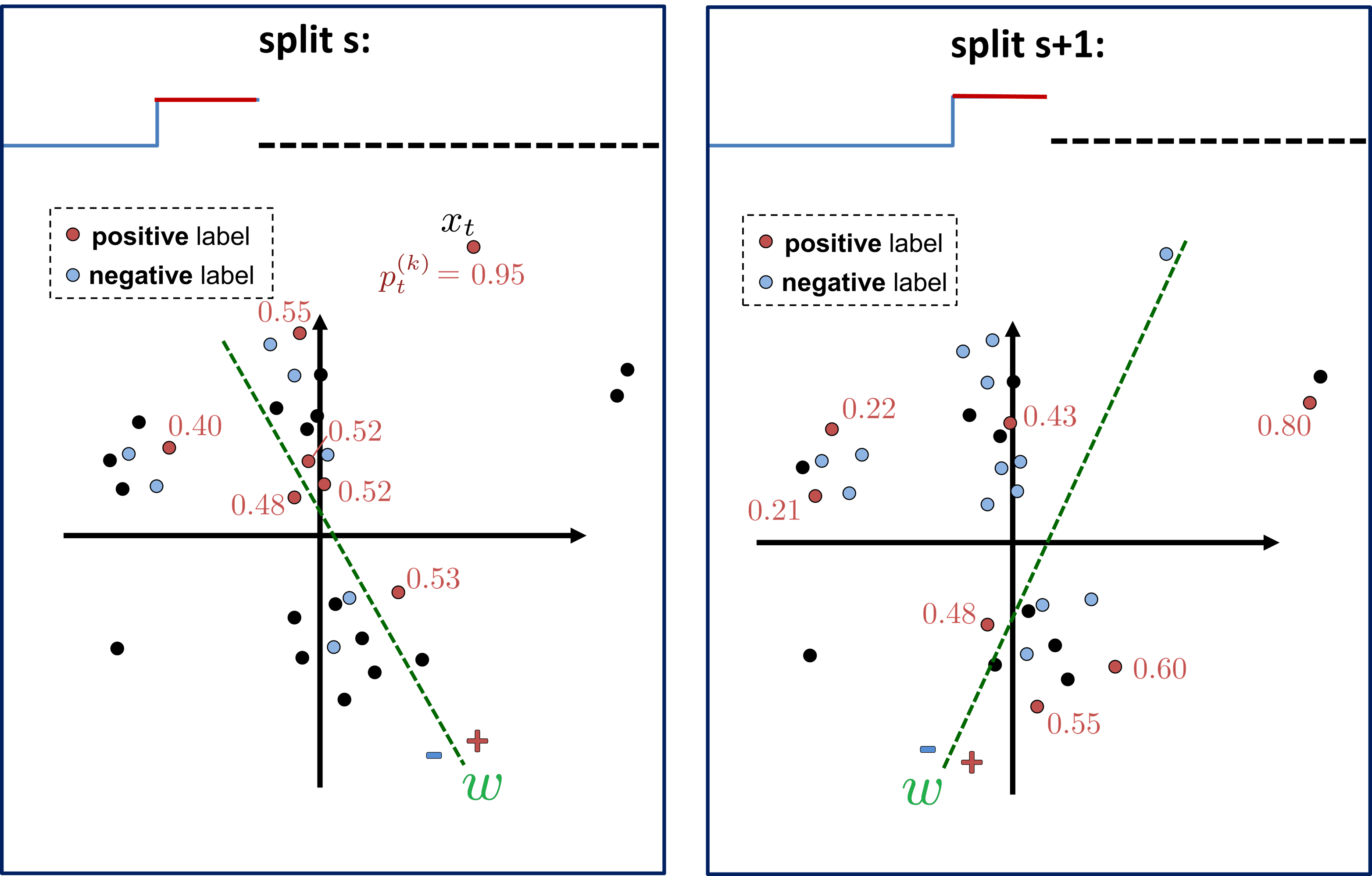}
    \caption{   
    \textbf{Visualizing two consecutive splits. } Points in red and blue have been labeled $1$ and $0$ respectively.  Here the window size $t_w=7$ (\# red points).  The values in red are the probability values $p_t$.  Familiars are close to the boundary ($p_t$ close to $0.5$) and therefore yield a low log odds score $a_t$.
} \label{fig:twosplits}
\end{minipage}
\hspace{2mm}
\begin{minipage}[t]{.35\textwidth}
    \centering
    \includegraphics[width=0.9\linewidth,height=33mm]{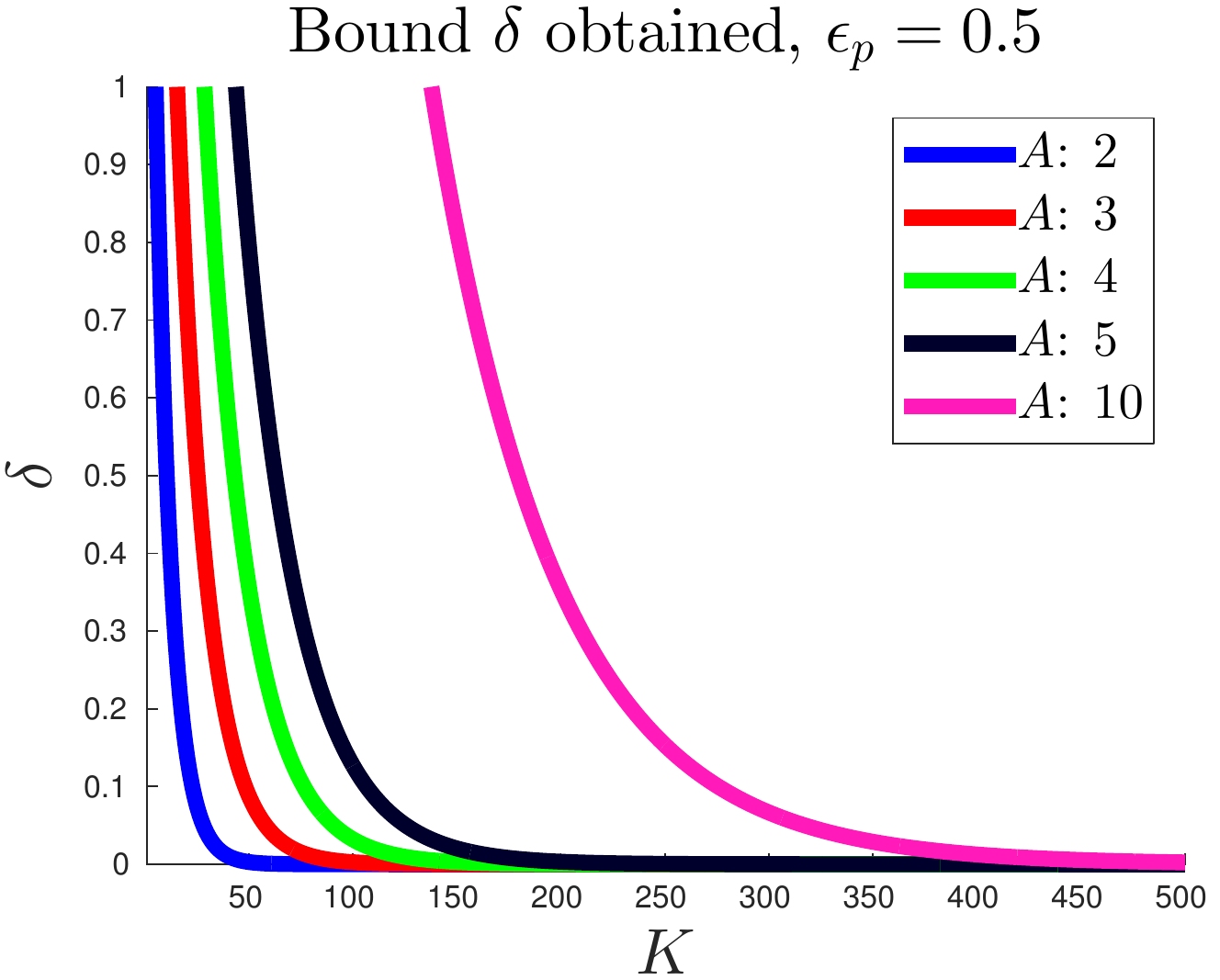}
    \caption{   Upper bound on the probability that any one of $A$ anomalies appears first in fewer than $1/A - \epsilon$ shuffles ($\epsilon=0.05$).  The resulting probability decreases exponentially with the number of shuffles.}
    \label{fig:theoryShuffles}    
\end{minipage}%
\end{figure}

\textbf{No shuffles - change detection.} If we were to remove permutation from our algorithm, it would perform simple change detection by testing the distinguishability of a sliding window of $t_w$ frames, where all frames before it are assumed to be familiar.  A conceptual example is shown in Figure~\ref{fig:twosplits}.

In the first iteration, a classifier $f$ is learned on the set of $2t_w$ points, where the first $t_w$ points are given the label 0 and the second $t_w$ points are given the label 1.  We call this set of labels a `split' of the data.  Each point $x$ that is labeled 1 is then given the score $f(x)$, which is the probability it belongs to class 1 instead of class 0 according to the classifier $f(x)$.  In our implementation, $f(x)$ is simply $ \frac{1}{1 + \exp{(-w^Tx)}}$ where $w$ minimizes the $l2$-regularized logistic loss.  We say the second set of frames are within a `sliding time window' or `sliding window', because in the next iteration these frames are reassigned a label of 0 and the next $t_w$ points are labeled 1\footnote{For simplicity, we describe the algorithm when the sliding window size and stride equal each other, $t_w$ = $\Delta t_w$. It is just as valid to shorten the stride to increase accuracy.  See Algorithm~\ref{algo:anomDetect}.}.  The process repeats until the sliding window reaches the end of the video.   As the sliding window reaches the end, events in the window are compared to all events in the past.  The higher $f(x)$ for a given point $x$, the larger the classifier's confidence that it can be distinguished from previous points. 

A sliding window is chosen rather than moving point-by-point for several reasons.  First of all, it provides inherent regularization for the classifier, since distinguishing any one point the rest can be misleadingly easy even if that point is familiar.  In addition, the number of splits the algorithm must compute is inversely proportional to the window size $t_w$.  It may seem that `polluting' the sliding window with familiars would ruin an anomaly's chance of being accurately scored.  However, this is not the case, as anomalies are more easily distinguished from the rest of the video, and therefore the chance that they fall near the resulting classifier boundary is low, while the probability that a familiar event does is high (see Figure~\ref{fig:twosplits} for intuition).

\textbf{Adding in shuffles - full anomaly detection.} As we pointed out earlier, the disadvantage of this approach without shuffles is the same as with other scanning techniques: temporal dependencies cause the algorithm to miss events that occur more than once and raise false alarms to events that may be prevalent later in the video but not in the beginning.  We therefore shuffle the order of the data and repeat the change detection process described to reduce the effect of the order.  Producing a series of distinguishability scores from classifiers learned on different permutations of frames can be thought of as testing a set of hypotheses.  If a series of classifiers are all able to easily distinguish a frame labeled ``1'' from many combinations of those labeled ``0'', it is likely an anomaly.

\textbf{Aggregating scores.} Once the scores have been computed for each shuffle, we average the results.  The average outperforms other methods of aggregation like the median and maximum.  After aggregating over shuffles, log-odds are computed as the final anomaly score. The full overview is explained in Algorithm~\ref{algo:anomDetect}.

\begin{algorithm}[t]
\textbf{Parameters}: $K$ (\#permutations), $t_w$ (window size), $\Delta t_w$ (window stride) \;
\KwIn{$\{x_1,..,x_T\}$ (descriptors for each frame)}
\KwOut{$\{a_1,...,a_T\}$ (estimates of anomalousness of each frame)}
Generate a random set of $K$ permutations $\{\sigma_1([1,...,T]),...,\sigma_K([1,...,T])\}$ \\
\For{ $k = 1,...,K$}{
    \For{all sliding windows $[t_\text{start},t_\text{start} + t_w)$ until $T$}{
        $y_{\sigma_k(t)} \gets
          \begin{cases}
           0    & \text{if } t < t_\text{start}\\
           1    & \text{if } t_\text{start} \leq t < (t_\text{start}+t_w)\\
          \end{cases}$ \\
        $w \gets$ TrainLogisticRegression($x_{\sigma_k}$,$y_{\sigma_k}$) \\ 
        $p_t^{(k)} \gets P(y_t^{(k)} = 1 | x_t,w), \> \forall (y_t^{(k)} == 1$) \\
    }
}
$\bar{p_t} \gets$ mean$(p_t^{(k)})$ across all $k$ \\
$a_t \gets \frac{\bar{p_t}}{(1-\bar{p_t})}$

\Return{$\{a_1,...,a_T\}$}
\caption{{\sc Anomaly Detection} Selects the most anomalous frames}
\label{algo:anomDetect}
\end{algorithm}

\section{Supporting Theory}
Our method is based on distinguishing two labeled subsets of data.  We consider here how to think about the tradeoffs between increasing and decreasing the window size and the number of shuffles given a choice of classifier.  We attempt only to provide intuition by considering simple analyses that suggest how one might understand the tradeoffs of parameters and classifier choices.

\textbf{Overall objective.} In order for the algorithm to work, the classifier needs to have enough capacity to be able to tell anomalies and familiars apart, but simple enough that it is unable to tell familiars apart from other familiars.  This capacity is a function of the classifier complexity, the number of points being compared (related to $t_w$), and the subset of points being compared (the shuffling).

\textbf{Classifier assumptions.} We must make basic assumptions about the classifier.  We assume that the classifier $f(x)$ is able to correctly distinguish between an anomaly and familiars.  In the case of logistic regression, for instance, we assume that $f$ will label an anomaly $1$ with relatively high confidence if it is learned on a set of familiars labeled 0 and an anomaly within a set of familiars labeled 1.  This means $f(x)=P(y=1|x,w)$ will be significantly higher than $0.5$ if $x$ is an anomaly and close to or below $0.5$ if $x$ is a familiar.

\textbf{Increasing the number of shuffles, K, removes temporal dependencies.} Here we show that the number of shuffles $K$ needed to remove temporal dependencies scales with the number of anomalies $A$ at $O(A \log A)$.

Suppose a video has $A$ anomalies and a significantly larger number of total frames, $A \ll T$. Consider the worst case scenario for our algorithm, where all $A$ anomalies are identical.  Formally, each of these anomalies have identical feature vectors: $x_{n_1}=x_{n_2}=...=x_{n_A}$.  This represents the worst case because if one such anomaly $x_{n_i}$ is labeled 0 while another anomaly  $x_{n_j}$ is labeled 1, the score of $f(x_{n_j})$ will be small.  In other words, an anomaly in the negative window can negate the score of another in the positive window.  Therefore, the best split of the data for a given anomaly instance is when it occurs first in the video.\footnote{ignoring the rare case when it occurs in the very first negative window.}  The goal of shuffling is to reorder the anomalies enough times that every anomaly gets the opportunity to be the first instance in the reordered set of frames.

More formally, let us define random variable $F$ as the event that an anomaly $x$ appears first in a given shuffle.  $F$ is a binary variable that occurs with probability $1/A$, so in expectation, the fraction of shuffles in which it appears first is also $\mu = 1/A$.  However, we would like to use as few shuffles as possible to ensure that the event $F$ gets close to its mean $1/A$.  This means we need a bound on the probability that $\bar{F}_K - \mu$ is less than some $\epsilon$ for $K$ shuffles.  A relative tolerance $\epsilon=(\epsilon_p/A)$ is required because as the desired fraction $1/A$ gets smaller, the effect of deviations $\epsilon$ on the score of an anomaly grows proportionally larger. For instance, $\epsilon_p=0.25$ requires that every anomaly is first in at least 75\% as many shuffles as it would be on average.
To obtain this bound, we apply the Relative Entropy Chernoff Bound~\cite{motwani2010randomized}\footnote{Kakade provides a quick summary for those unfamiliar with these bounds~\cite{boundslecture}: \url{http://stat.wharton.upenn.edu/~skakade/courses/stat928/lectures/lecture06.pdf}}.

The Chernoff bound for Bernoulli random variables states that the average $\bar{F}_K$ of $K$ variables with mean $\mu$ falls below $\mu - \epsilon$ for some $\epsilon > 0$ with probability:
\begin{equation}\label{eqn:1}
\Pr\left(\bar{F}_K \le \mu - \epsilon\right) \le e^{-K  \text{ KL}(\mu-\epsilon,\mu)},
\end{equation}
where KL$(p_1,p_2)$ is the KL divergence between two Bernoulli distributions with parameters $p_1$ and $p_2$.  This is a well-known formula, and the KL divergence KL$(\mu-\epsilon,\mu)$ with $\mu=1/A$, $\epsilon=\epsilon_p/A$ is:
\begin{equation}\label{eqn:3}
\text{KL}(\mu-\epsilon,\mu)=\frac{1}{A}\Big[\log(1-\epsilon_p)(1-\epsilon_p) + \log\Big(\frac{\epsilon_p}{A-1}\Big)(A-1+\epsilon_p)\Big]
\end{equation}
With Equations~\ref{eqn:1} and~\ref{eqn:3}, we have bounded the probability that a single anomaly will not appear first in a large enough fraction of shuffles.    To extend this to all $A$ anomalies, we apply the union bound to all events $\bar{F}_K^{(i)}$ to get our final bound:
\begin{equation}\label{eqn:4}
\delta:=\Pr\left(\bigcup_{i=1}^A \Big(\bar{F}_K^{(i)} \le \mu - \epsilon\Big)\right) \le A e^{-K  \text{ KL}(\mu-\epsilon,\mu)}
\end{equation}

The bounding probability $\delta$ is defined in terms of $A$, $K$, and $\epsilon_p$.  Values of this bound for different values of $K$, $\delta$, and $A$ are depicted in Figure~\ref{fig:theoryShuffles}.  We are interested in choosing the number of shuffles $K$ for a given $\delta$, $A$, and $\epsilon_p$, so we solve Equation~\ref{eqn:4} for $K$:

\begin{equation}\label{eqn:5}
K \ge \frac{\log(\frac{A}{\delta})}{\text{KL}(\mu-\epsilon,\mu)}
\end{equation}

In big O terms, for fixed $\delta$ and $\epsilon_p$, we need $O(A \log A)$ shuffles to reorder the anomalies enough times to equally score them.



\textbf{Larger window sizes decrease the effect of overfitting classifiers.}\label{sec:theorycomplexity} Given a choice $K$, there is one more parameter in the framework that requires care: the window size $t_w$.  Increasing the window stride decreases the computational load on the system (more splits per shuffle). In terms of performance, it may also seem best to make $t_w$ as small as possible because anomalies are easier to distinguish in a smaller window size. In addition, with large window sizes, anomalies of different types can fall within the same window and `interfere' with each others' scores. However, decreasing the window size beyond a certain point also reduces performance, as the classifier overfits and familiars become distinguishable from other familiars.  In other words, we must choose $t_w$ to be able to trade off our ability to distinguish anomalies without being able to distinguish familiars.  We consider here a theoretical sketch explicating the relation between the complexity of the classifier and the choice of window size $t_w$. 

Assume we have computed a complexity metric \cite{ben2010theory} for our chosen classifier $f$.  In this instance, we will work with the Rademacher complexity\footnote{The insights that follow generalize to VC dimension and other complexity measures.} $\mathcal{R}(m)$, where $m$ is the size of the subset of points being classified~\cite{bartlett2003rademacher}.  A higher Rademacher complexity indicates the classifier is able to more easily distinguish randomly labeled data.  For instance, a highly regularized linear classifier has a much lower Rademacher complexity than a RBF-kernel SVM or complex neural network.  This metric is especially convenient because it is measured relative to the data distribution (so it adapts to the video) and can be empirically estimated by simply computing a statistic over randomly labeled subsampled data\cite{bartlett2003rademacher}\footnote{For a useful introduction, see~\cite{ninalecture}.}.

Given that we have a classifier of complexity $\mathcal{R}$, we can adjust the window size to decrease the probability that familiars can be distinguished from each other.  Generalization bounds provide a way to relate the error from overfitting or noisy labels to the classifier complexity and dataset size.  In our case, the true error, is the error from classifying anomalous or familiar points incorrectly according to their true labels.  The training error is the error when we trained the classifier on our synthetic labels.

While a careful analysis requires understanding errors in the \textit{fixed design} \cite{borenstein2010basic} setting, the traditional \textit{i.i.d.} random design provides crude guidance on algorithm behavior and trade-offs.  In this setting, the Rademacher complexity provides us a generalization bound~\cite{bartlett2003rademacher}. For i.i.d. samples, the difference between the estimated and true error in classifying a set of $m$ datapoints will be ${\text{err} - \widehat{\text{err}} \le \mathcal{R}(m) + \mathcal{O}\bigg(\sqrt{\log \frac{1 /\delta}{m}}\bigg)}$
with probability $1 - \delta$.  This gives a good intuition for how the overfitting of our classifier relates to its complexity and the number of points $m$ that we train on.  This bound follows our intuition: (1) as $m$ decreases, the chance of classifying incorrectly increases; (2) as $\mathcal{R}(m)$ increases, the classifier complexity increases and we have a better chance of incorrectly distinguishing familiars from other familiars.

Due to this intuition, we choose a simple classifier, $l2$-regularized logistic regression.  Window size $t_w$ is most easily chosen through empirical testing; if the variance in the anomaly signal is large, familiars are too easy to tell apart and the window size $t_w$ should be decreased.  When no anomalies are visible, $t_w$ should be increased.


\section{Experiments}
\textbf{Dataset.} We tested our algorithm on the Avenue Dataset~\cite{lu2013abnormal}\footnote{\url{http://www.cse.cuhk.edu.hk/leojia/projects/detectabnormal/dataset.html}} as well as the Subway surveillance dataset~\cite{adam2008robust}, the Personal Vacation Dataset~\cite{ito2012}, and the UMN Unusual Activity Dataset~\cite{UMN}.

The Avenue dataset contains 16 training videos and 21 testing videos, and locations of anomalies are marked in ground truth pixel-level masks for each frame in the testing videos. The videos include a total of 15324 frames for testing.  Our algorithm was permitted to use only the testing videos, and performed anomaly detection with no assumptions on the normality of any section of this video.  This is in stark contrast to other methods, which must train a model on a set of frames from the training videos and/or from pre-marked sections of video.  There are several other datasets available for anomaly detection, and our algorithm demonstrated reasonable success on all of the ones we tested\footnote{See supplementary material for more results, including results on individual videos in the UMN and Avenue datasets.}. We focus on the Avenue Dataset specifically because it was more challenging than staged datasets (such as the UMN Unusual Activity Dataset) and is more recent with more specific labeling than others, such as the Personal Vacation dataset.  The dataset is also valuable because the method in~\cite{lu2013abnormal} has publicly available code and results, so we were able to compare with the same implementation and features as a recent standard in anomaly detection.  The UCSD pedestrian anomaly detection dataset~\cite{mehran2009abnormal} is another well-labeled and recent dataset, but nearly half of the frames in each test video contain anomalies, so the provided anomaly labels are not applicable in our unsupervised setting.  More precisely, in our setting, no frames would be defined as anomalous since the activities labeled as such in the dataset often compose half of the video.\footnote{Other videos are unusable for a similar reason; for instance, 2 of the 8 videos available from~\cite{YorkVideos} contain more than 50\% frames with at least one anomaly, and only 2 contain fewer than 20\% anomalous frames (one of which is the Subway exit sequence).}

\textbf{Implementation.} For the Avenue dataset, we follow the same feature generation procedure as Lu~\cite{lu2013abnormal}, courtesy of code they provided upon request.  The features computed on the video match their method exactly, resulting in gradient-based features for 10x10x5 (rows x columns x frames) spatiotemporal subunits in the video.  After PCA and normalization, each subunit is represented by a 100 dimensional vector.  Using the code provided by the authors, we are able to run their algorithm alongside ours on the same set of features.  Following their evaluation, we treat each subunit as a `frame' in our framework, classifying each subunit independently.  The results are smoothed with the same filter as~\cite{lu2013abnormal}.

We used liblinear's $l2$-logistic regression for the classifier $f$ in the framework.  We experimented with several values of $\lambda$ across other videos and found that as long as the features are whitened, $\lambda$ within an order of magnitude of $1$ gives reasonable results (see Table~\ref{table:table1}.  We only needed $10$ shuffles to get adequate performance, likely because there were few anomalies per video.  We display our results for a window size of $10$.  The algorithm is trivially parallelizable across shuffles and splits; we provide a multithreaded version that runs splits simultaneously across all allocated CPUs.  This ability to increase accuracy by operating more units in parallel is a significant benefit of this framework.  In contrast, scanning techniques that update models in an online fashion must operate serially; their computation cost is dependent on the desired accuracy.  An implementation of our method is available online.\footnote{\url{http://www.cs.cmu.edu/~adelgior/anomalyframework.html}}

\textbf{Evaluation metric.} We are interested in identifying every instance of an anomaly.  We also care about proposing frames for a human to review, meaning we would like to evaluate the fidelity of the anomalousness with a human rating of anomalousness.  Consequently, we avoid metrics that score anomaly detections in an event-detection style, where flagging a single frame is counted as a successful detection of adjacent frames. In addition, metrics like Equal Error Rate (EER) can be misleading in the anomaly detection setting.\footnote{Consider the case when only 1\% of the video is anomalous: the EER on an algorithm that markes all frames normal would be 1\%, outperforming most modern algorithms.  This extreme class imbalance is less prevalent in current standard datasets, but will become an apparent problem as more realistic datasets become prevalent.}  Therefore, we using ROC curves and the corresponding area under the curve (AUC) as the evaluation metric, computed with reference to human-labeled frame and pixel ground truth.

\begin{figure*}[t]
\centering
\begin{subfigure}{.33\textwidth}
  \centering
  \includegraphics[width=.7\linewidth]{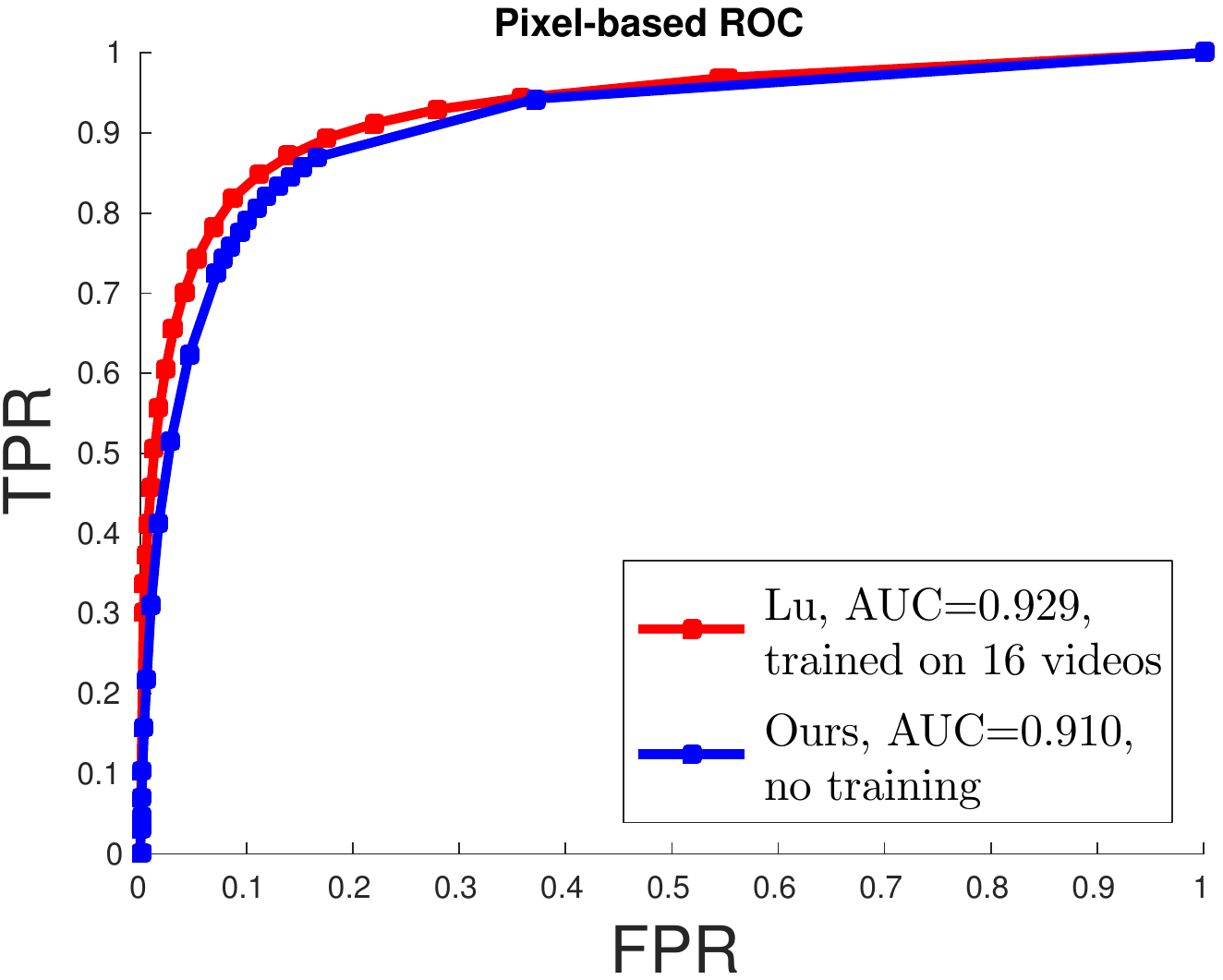}
  \caption{   Pixel-based ROC on Avenue dataset}
  \label{fig:ROC_avenue_pixel}
\end{subfigure}%
\begin{subfigure}{.33\textwidth}
  \centering
  \includegraphics[width=.7\linewidth]{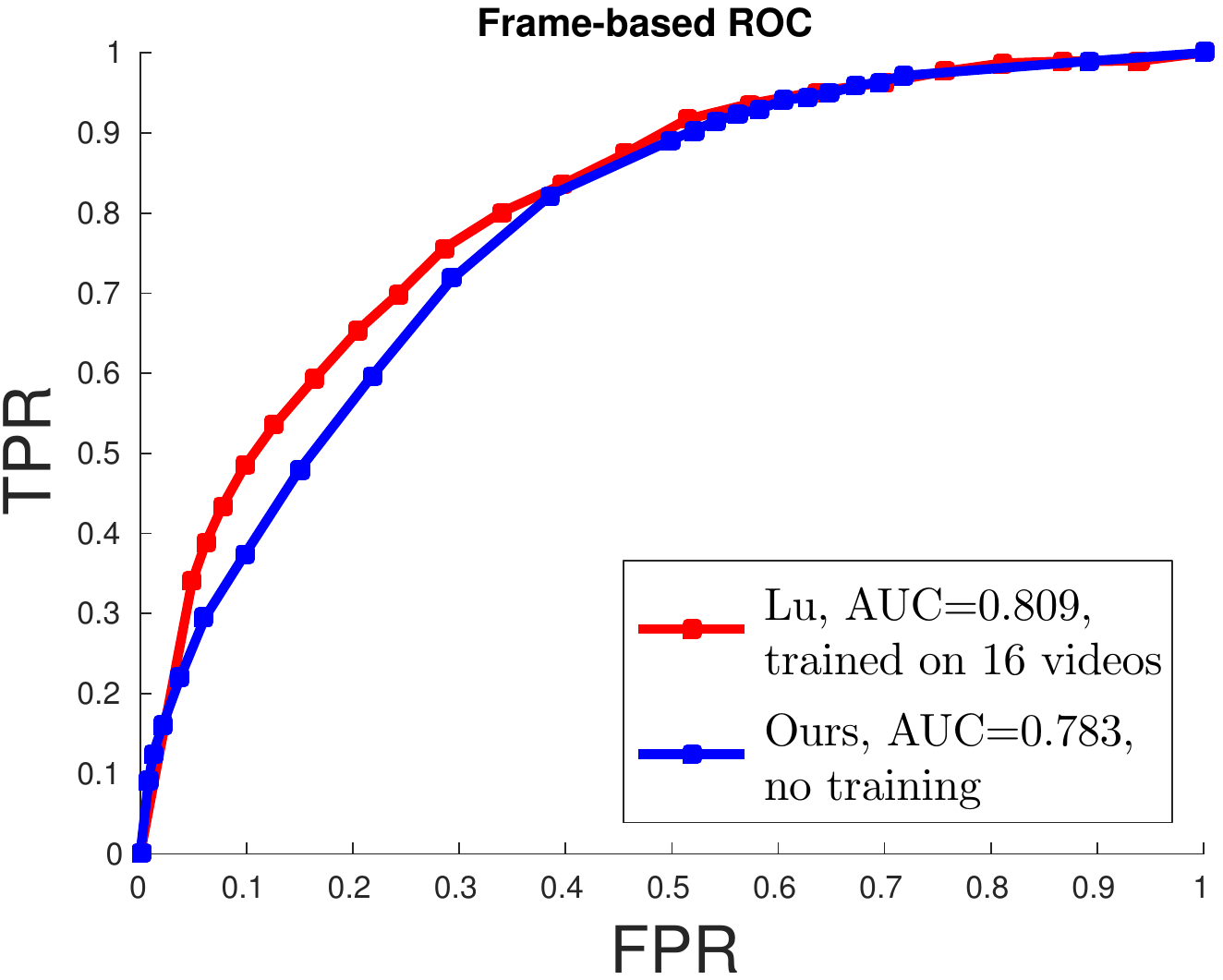}
  \caption{   Frame-based ROC on Avenue dataset}
  \label{fig:ROC_avenue_frame}
\end{subfigure}
\begin{subfigure}{.33\textwidth}
  \centering
  \includegraphics[width=.8\linewidth]{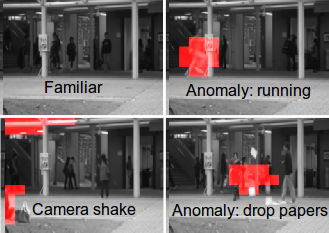}
  \caption{   Pixel-based detection examples}
  \label{fig:detections}
\end{subfigure}
\caption{   Performance on the Avenue Dataset.  ROC curves in (a) and (b) show our performance nearly matches this algorithm, while we require no training data.  Detection examples shown in (c) show a correctly classified familiar frame, two detected anomalous frames (running and dropping papers), and a false alarm (camera shake).}
\label{fig:ROC_avenue}
\end{figure*}

\textbf{Results.} Figure~\ref{fig:ROC_avenue} shows example detections and the resulting ROC curves and AUC values for our algorithm and~\cite{lu2013abnormal} on the Avenue dataset.  Note our algorithm operates in a separate setting -- it does not (a) use any sequence other than an individual test video, (b) obtain a guiding form for models of familiarity, or (c) assume any partition of the video is familiar. Even with these additional challenges, we are able to obtain near-state-of-the-art performance.

The ROC curve sheds some light on the possible performance bottleneck for these algorithms: the last half of the curves for the two algorithms match closely. This highlights that for difficult anomaly instances, a similar number of false positives seem to be commonly detected by both algorithms. We believe that both are hitting a limitation with the encoding of events in the feature space: either the feature space is not descriptive enough or several other instances appear as anomalous as the true anomalies in feature space.  Example detections and per-video analysis (see Supplementary) also shed light on our method's behavior.  Since our algorithm is operating only on the test sequence, it exhibits false positives such as the only time someone enters the foreground from the right. This is penalized because the provided ground truth was marked relative to the training data.  In addition, by using features that operate on 15-frame chunks of time, we often detect events as early as 15 frames too soon.

In Table~\ref{table:table1} we show that our algorithm's robust performance across a range of parameters on the Avenue dataset. These results show that for sub-optimal parameter choices (common in the unsupervised learning setting), shuffles can improve performance. Imagine an under-regularized classifier ($\lambda$ too small). The classifier will too easily distinguish normal points from subsets of the data, but this effect is reduced as the number of shuffles increases. A similar argument follows for other sub-optimal parameters. The major benefits of shuffling cannot be seen in the commonly used datasets for anomaly detection because the test sequences are too short to show context changes or multiple instances of the same anomaly that are more commonly found in real-world scenarios.

In addition, we report our AUC values on the Subway dataset (exit: 0.8236; entrance: 0.6913). Results for one of the videos in the Personal Vacation Dataset are shown in Figure~\ref{fig:seaplots}. Detailed UMN Dataset results can be found in the supplementary material.  Our method outperforms \cite{lu2013abnormal} on all but one scene.  While the AUC values are good (average=0.91), the average AUC for both our method and the sparse method is still lower than the sparse and model-based approaches reported in~\cite{li2014anomaly}; this indicates that a change in features could make up for the difference in the performance gap.

\begin{minipage}{.48\linewidth}
    \centering
    \captionof{table}{Frame-based AUC averaged across all 21 test videos of the Avenue dataset.  All instances of $K=10$ outperform their $K=0$ counterparts.}
     \tiny
    \begin{tabular}{@{}rrrrcrrrcrrr@{}} \toprule
            & \multicolumn{2}{c}{$K=0$} & \phantom{abc}& \multicolumn{3}{c}{$K=10$}\\
            \cmidrule{2-4} \cmidrule{6-8}
            $t_w$& $10$ & $100$ & $1000$&& $10$ & $100$ & $1000$ \\ \midrule
            $\lambda=0.01$ & 0.8697 & 0.8149 & 0.7731 && 0.8950 & 0.8951 & 0.8085\\
                $\lambda=1$ & 0.8921 & 0.8736 & 0.7731 && 0.8957 & 0.8954 & 0.8085\\
            $\lambda=100$ & 0.8947 & 0.8922 & 0.7751 && 0.8958 & 0.8953 & 0.8088\\
        \bottomrule
    \label{table:table1}
    \end{tabular}
\end{minipage}
\hspace{1mm}
\begin{minipage}{.48\linewidth}
\centering
\includegraphics[width=.8\linewidth]{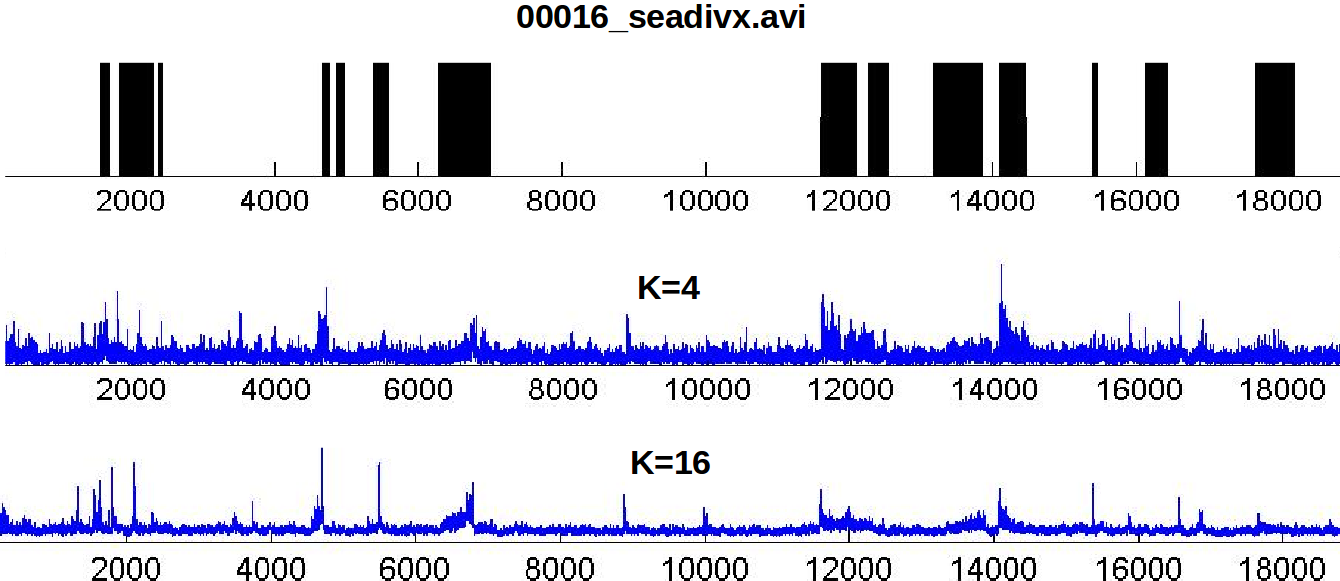}
\captionof{figure}{Results on video 00016 in the Personal Vacation Dataset.  As the number of shuffles increases, the signal-to-noise ratio increases.}
\label{fig:seaplots}
\end{minipage}

\section{Discussion}
We have developed a method for identifying anomalies in videos in a setting that is independent of the order in which anomalies appear and requires no separate training sequences.  The permutation-testing methodology requires no assumptions about the content in the descriptors of each frame and the user is able to ``plug in'' the latest optimal features for their video as long as the anomalous frames are distinguishable in this space.  No training data needs to be collected or labeled within the test video.  We show that our anomaly detection algorithm is able to perform as well as state of the art on standard datasets, even when we adjust the setting by removing training data.  The lack of assumptions on the content or location of familiars is valuable for finding true anomalies that had previously remained unseen.

\textbf{Acknowledgements.} This research was supported through the DoD National Defense Science \& Engineering Graduate Fellowship (NDSEG) Program and NSF grant IIS1227495.




\bibliographystyle{splncs} 
\bibliography{mybib}

\end{document}